%% file: NormDetect.tex
\documentclass[sigconf,nonacm,balance=false]{acmart}
\setcopyright{none}
\acmConference[WWW '27]{The ACM Web Conference 2027}{May 10--14, 2027}{Dublin, Ireland}
\acmYear{2027}
\usepackage{booktabs}
\usepackage{multirow}
\usepackage{pifont}
\usepackage{xspace}
\usepackage{amsmath}
\usepackage{graphicx}
\usepackage{subcaption}
\usepackage{algorithm}
\usepackage{algpseudocode}
\usepackage{colortbl}
\usepackage{placeins}
\newcommand{\sys}{NormDetect\xspace}
\newcommand{\shatter}{token shattering\xspace}

\newcommand{\xmark}{\ding{55}}

\begin{document}

\title{NormDetect: Anchored Calibration for Low-False-Positive Detection of Machine-Generated Web Text}

\author{Hanshuai Cui}
\affiliation{\institution{School of Artificial Intelligence, Beijing Normal University, Beijing 100875, China}\city{}\country{}}
\affiliation{\institution{Institute of Artificial Intelligence and Future Networks, Beijing Normal University, Zhuhai 519087, China}\city{}\country{}}

\author{Fanshuai Meng}
\affiliation{\institution{Institute of Artificial Intelligence and Future Networks, Beijing Normal University, Zhuhai 519087, China}\city{}\country{}}

\author{Zhiqing Tang}
\affiliation{\institution{Institute of Artificial Intelligence and Future Networks, Beijing Normal University, Zhuhai 519087, China}\city{}\country{}}

\begin{abstract}
Web platforms screen posts for text written by large language models, and every false flag hits a human writer.
Detectors must therefore run at a very low false positive rate (FPR).
Zero-shot detectors such as Binoculars score text with frozen language models, but invisible edits such as homoglyphs shatter tokenization and push machine text into the human range.
Unicode normalization undoes these edits, yet the normalized score still misses half of the machine text at a strict FPR, under attacks that leave nothing to clean.
We find that with the Falcon-7B scoring pair, heads on token statistics recall less than the normalized score at this FPR, and several flag too many human posts in some sections.
We propose \sys, which anchors a learned per-section head to the normalized zero-shot score.
A blend weight chosen on training folds under a human FPR cap combines the two scores, and one threshold is calibrated on human text only.
On eight sections of the RAID benchmark with eleven attacks, \sys raises recall at a 0.1\% calibrated FPR by 5.0 points over normalized Binoculars.
Its human FPR bound stays at most 1\% in every section, and no other detector within this bound recalls as much.
\end{abstract}

\begin{CCSXML}
<ccs2012>
<concept><concept_id>10002951.10003260</concept_id><concept_desc>Information systems~World Wide Web</concept_desc><concept_significance>500</concept_significance></concept>
<concept><concept_id>10010147.10010178.10010179</concept_id><concept_desc>Computing methodologies~Natural language processing</concept_desc><concept_significance>300</concept_significance></concept>
</ccs2012>
\end{CCSXML}
\ccsdesc[500]{Information systems~World Wide Web}
\ccsdesc[300]{Computing methodologies~Natural language processing}
\keywords{machine-generated text detection, Web content analysis, false positive rate, Unicode normalization, calibration}

\maketitle

%======================================================================
\section{Introduction}
\label{sec:intro}
%======================================================================
Review sites, forums, question-answering boards and news comment sections now receive text written by large language models (LLMs).
Fake opinion posts have long been studied and detected~\cite{jindal2008,ott2011}, and measurements find LLM-written text in social media posts, consumer and corporate writing, scientific abstracts and peer reviews~\cite{sun2025aigt,liang2025society,kobak2025,liang2024peer}.
About 35\% of new websites published by mid-2025 are classified as generated or assisted by artificial intelligence~\cite{dolezal2026}.
Platforms therefore screen posts for this content, and companies and governments discuss labelling it~\cite{raid}.
This screening detects machine-generated text on the Web, a topic of the Web Mining, Multimedia and Multilingual Content Analysis track.
Its costly error is the false positive, because a flagged human writer may lose a review, an account or trust in the platform.
A platform therefore runs a detector at a very low false positive rate (FPR) on human text, such as 0.1\%.
That rate must hold in every section of the site, a check we call the per-section guard.
Under this constraint every point of recall counts, since each missed machine post stays online.

\begin{figure}[t]
\centering
\includegraphics[width=\linewidth]{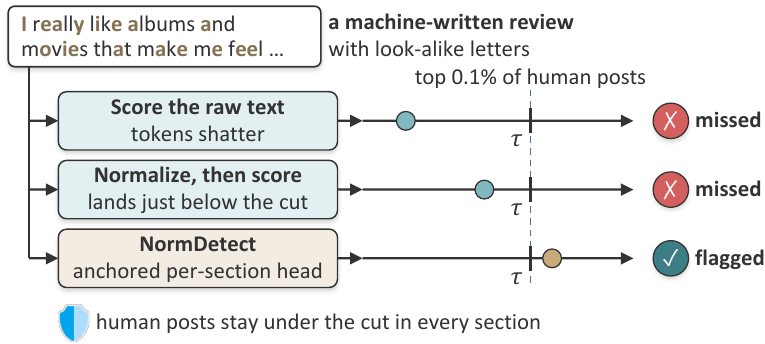}
\Description{A machine-written review with look-alike letters goes through three detectors drawn as score axes with a threshold. Scoring the raw text and scoring the normalized text leave it below the threshold, and NormDetect flags it.}
\caption{A machine-written review with look-alike letters at a 0.1\% calibrated FPR. Raw and normalized Binoculars score it below $\tau$, and \sys flags it.}
\label{fig:teaser}
\end{figure}

\begin{figure*}[t]
\centering
\includegraphics[width=\textwidth]{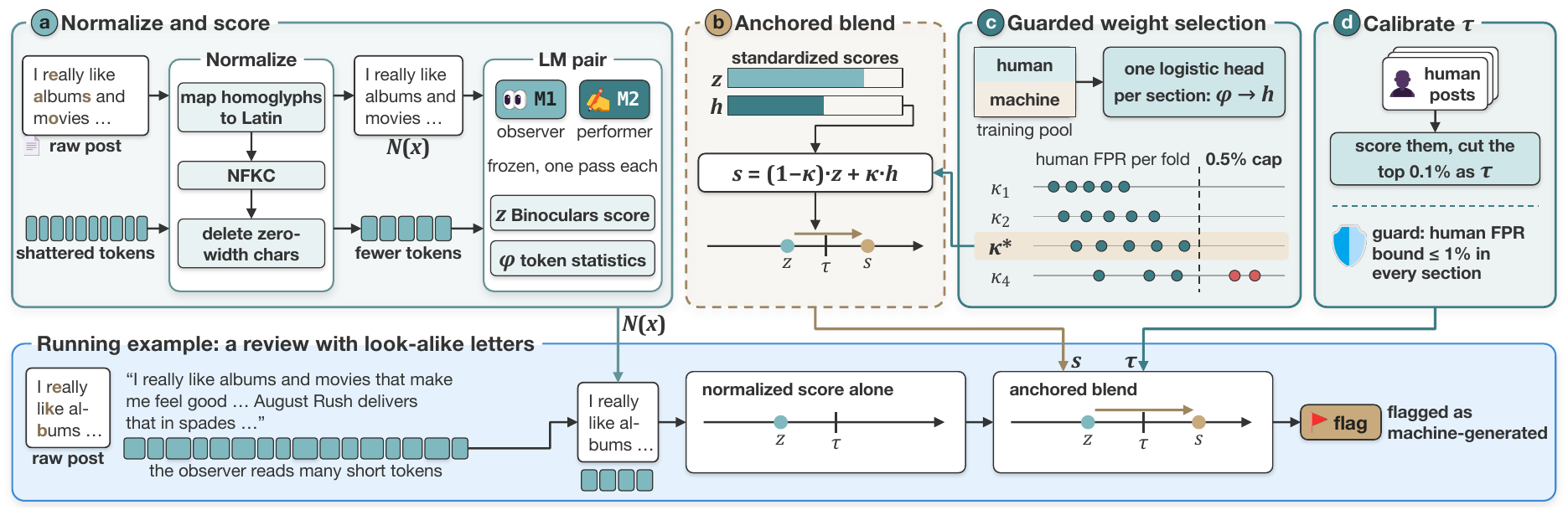}
\Description{Four-panel pipeline: text normalization and two frozen scoring models, an anchored blend of a per-section head with the normalized Binoculars score, an offline choice of the blend weight, which keeps the best weight whose human false positive rate stays at most 0.5\% on all five training folds, and a calibrated threshold.}
\caption{Overview of \sys. A per-section head is anchored to the normalized Binoculars score $z$, with $\kappa^*$ set on training folds and $\tau$ on human calibration text.}
\label{fig:overview}
\end{figure*}

Detectors fall into three families.
Zero-shot detectors score a document with frozen language models and need no training.
They use the likelihood or rank of the observed tokens~\cite{gltr,detectllm}, the curvature of the likelihood~\cite{detectgpt,fastdetectgpt}, the ratio of perplexity to cross-perplexity of two models~\cite{binoculars}, regenerated continuations~\cite{dnagpt}, or finer token statistics~\cite{lastde,telescope}.
Robust pipelines clean the input with Unicode normalization against character edits~\cite{conpara,humanizing} or clip token contributions against outliers~\cite{clippedlrt}.
Trained detectors fit a classifier on labelled text, from fine-tuned encoders~\cite{radar} to heads on surprisal or likelihood features~\cite{diveye,biscope}.

However, these detectors lose much of their recall at a low FPR.
They overlook what character edits do to the tokenizer and what a learned head does to the tail of human scores.
We find three effects on the RAID benchmark~\cite{raid} at a 0.1\% calibrated FPR (Section~\ref{sec:motiv}).
First, homoglyphs and zero-width spaces leave a post unchanged to a reader but split it into 1.5--1.8 times as many tokens.
The recall of Binoculars drops from 75.5\% to 2.0--4.2\% on such posts.
We call this \shatter (Figure~\ref{fig:teaser}).
Second, normalization repairs these posts without flagging a single extra post in a held-out set of human posts.
Yet the normalized score still misses half of the machine text, under attacks such as paraphrasing that leave nothing to clean.
Third, with the Falcon-7B pair of Binoculars, heads trained on token statistics recall less than the normalized score at this FPR.
Several of them also flag too many human posts in some sections.

Building a better detector on these findings is hard in two ways.
\textbf{(C1)} \emph{How can a learned head add recall at a 0.1\% FPR?}
On the Falcon-7B pair, a head ranks the tail of human text worse than one zero-shot score.
Replacing the score with the head gives up 3--27 points of recall, and more features make the tail worse.
\textbf{(C2)} \emph{How much should the detector trust the head?}
The weight must be set without evaluation data, and every section must keep its human FPR.
Tuning it on development data would leak, and a weight chosen for recall alone has no check on the human tail of each section.

Our detector, \sys (normalize, then detect), anchors a learned head to the normalized zero-shot score (Figure~\ref{fig:overview}).
First, to address C1, \sys normalizes each document and computes the Binoculars score and eight token statistics in the same two forward passes.
It blends the standardized score with the output of a small head for the document's section.
Second, for C2, it chooses the blend weight on training folds under a cap on human FPR.
One global threshold is then calibrated on human text alone.
On eight RAID sections with eleven attacks, \sys recalls 55.5\% at a 0.1\% calibrated FPR (Section~\ref{sec:main}).
This is 5.0 points above normalized Binoculars and 2.3 points above the strongest prior detector that holds the per-section guard.
With the Gemma-4-12B pair, it recalls more than three times as much as any baseline on that pair.
We make the following contributions.
\begin{itemize}
\item We identify \shatter and show that normalization repairs it but leaves half of the machine text undetected. With the Falcon-7B pair, heads on token statistics also recall less than the normalized score at a 0.1\% FPR.
\item We propose \sys, which blends a per-section head with the normalized Binoculars score and selects the blend weight under a human FPR cap without evaluation data.
\item Experiments against seventeen detectors on eight Web sections and eleven attacks show that \sys is the only trained detector that holds the guard and beats normalized Binoculars in every section and on redrawn splits.
\end{itemize}

%======================================================================
\section{Related Work}
\label{sec:related}
%======================================================================
\begin{figure*}[t]
\centering
\begin{subfigure}[t]{0.354\linewidth}\includegraphics[width=\linewidth]{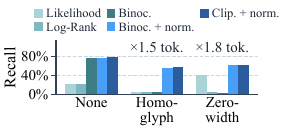}\caption{Character edits}\end{subfigure}%
\begin{subfigure}[t]{0.313\linewidth}\includegraphics[width=\linewidth]{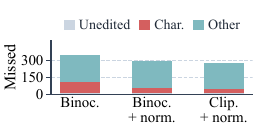}\caption{Remaining misses}\end{subfigure}%
\begin{subfigure}[t]{0.322\linewidth}\includegraphics[width=\linewidth]{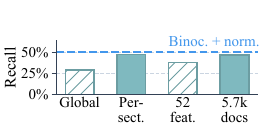}\caption{Heads alone}\end{subfigure}
\Description{Three bar charts. Recall of five zero-shot detectors on unedited, homoglyph and zero-width documents; machine documents missed by three detectors split by attack type; recall of four learned heads against the normalized Binoculars score.}
\caption{Measured motivation on RAID with the Falcon-7B pair at a 0.1\% calibrated FPR.}
\label{fig:motiv}
\end{figure*}

\paragraph{Zero-shot detection.}
Zero-shot detectors score text with frozen language models.
GLTR~\cite{gltr} showed that machine text favours high-likelihood, low-rank tokens, and DetectLLM~\cite{detectllm} combined both into the log-likelihood log-rank ratio (LRR).
DetectGPT~\cite{detectgpt} measured the curvature of the likelihood under perturbations, and Fast-DetectGPT~\cite{fastdetectgpt} replaced the perturbations with a sampling step that has an analytic solution.
Binoculars~\cite{binoculars} divides the perplexity of a text by its cross-perplexity between two related models.
DNA-GPT~\cite{dnagpt} compares a text with regenerated continuations, Glimpse~\cite{glimpse} lets these methods use proprietary models, and Lastde~\cite{lastde} and Telescope~\cite{telescope} read finer patterns of the token probability sequence.
Luminol-AIDetect~\cite{luminol} compares the perplexity of a text before and after shuffling its sentences or words, and classifies the shift with class densities fitted per domain.
These scores run on the tokens they receive, so an edit that changes tokenization changes the score.
\sys keeps such a score as its anchor and adds a learned correction on top of it.

\paragraph{Robustness to edits.}
Benchmarks such as RAID~\cite{raid} and DetectRL~\cite{detectrl} test detectors under paraphrasing, misspelling and character attacks, and MAGE~\cite{mage} and M4~\cite{m4} test unseen domains and generators.
Paraphrasers evade many detectors~\cite{dipper}, invisible and look-alike characters fool text classifiers~\cite{badchars,textbugger}, and homoglyphs fool machine-text detectors~\cite{wolff2020,homoglyphattack}.
Other work removes these characters with Unicode normalization before a supervised detector~\cite{conpara} or asks an LLM to map homoglyphs back to Latin letters~\cite{lies2025}.
Unicode compatibility normalization (NFKC) alone misses cross-script homoglyphs that an explicit map restores~\cite{humanizing}.
Clipped likelihood-ratio tests bound the contribution of outlier tokens~\cite{clippedlrt}.
Instruction-tuned models still read text in non-standard tokens~\cite{brokentokens}.
Byte-level models avoid subword tokenizers altogether~\cite{blt}, but the scoring pairs of zero-shot detectors read subword tokens.
These defences repair the input or bound the score.
\sys uses normalization as given and adds learned evidence for the machine text that normalization leaves undetected.

\paragraph{Trained detectors and low FPR}
Trained detectors fine-tune encoders~\cite{solaiman2019,radar,debertav3} or fit classifiers on likelihood and surprisal features~\cite{ghostbuster,diveye,biscope,adadetectgpt}, and surveys cover the wider field~\cite{survey}.
Watermarks~\cite{watermark,synthid} solve a different problem, since they require control over the generator.
Detectors flag human text unevenly, for example the writing of non-native speakers~\cite{liang2023bias,li2025roles}, so a deployed detector must bound its false positives on every kind of human text.
Classification under a false positive constraint is a Neyman--Pearson problem~\cite{neyman1933,nptong}, and benchmarks that fix the FPR report recall at a 1--5\% FPR~\cite{raid,tufts2025}.
MCP sets one conformal threshold per text-length bin to bound the FPR of a zero-shot score~\cite{mcp}.
It moves the threshold, while \sys changes the score that the threshold cuts.
\sys calibrates at 0.1\% and checks the bound in every section of the site.

%======================================================================
\section{NormDetect}
\label{sec:method}
%======================================================================

\subsection{Preliminaries}
\label{sec:pre}

\paragraph{Task.}
A Web platform receives documents $x$ posted to known sections $d(x)\in\mathcal{D}$, such as news, reviews, recipes or a Reddit-style forum.
A detector assigns each document a score $s(x)$ and flags it as machine-generated when $s(x)\ge\tau$.
Flagging a human writer is the costly error, so the platform fixes a small target false positive rate (FPR) $\alpha$ on human text.
We use $\alpha=0.1\%$ throughout.
Recall is the true positive rate (TPR) on machine-generated documents at the threshold $\tau$.

\paragraph{Calibration and guard.}
The threshold $\tau$ is set once on a pool $\mathcal{H}_{\text{cal}}$ of human documents, so that a share $\alpha$ of them scores above it.
The detector never sees labels of the documents it is later judged on.
A threshold that meets $\alpha$ on the pooled calibration set can still flag far more human text in one section.
We therefore check every detector on a separate human set $\mathcal{H}_{\text{guard}}$.
A detector \emph{holds the guard} when, in every section and in the pooled set, the one-sided 95\% Clopper--Pearson upper bound~\cite{clopper1934} of its human FPR is at most 1\%.

\paragraph{Binoculars score.}
Binoculars~\cite{binoculars} scores a document with two language models that share a tokenizer, an observer $M_1$ and a performer $M_2$.
For a document tokenized as $x_{1:L}$, the score is the ratio:
\begin{equation}
B(x)=\frac{\log \mathrm{PPL}_{M_2}(x)}{\log \mathrm{X\text{-}PPL}_{M_1,M_2}(x)},
\label{eq:bino}
\end{equation}
where the numerator is the mean negative log-likelihood of the observed tokens under the performer, as in the released code of Binoculars.
The denominator is the mean cross-entropy between the next-token distributions of the two models.
Machine text has a low ratio.
We write $z(x)=1-B(x)$, so that a higher score means machine text for every detector in this paper.

\paragraph{Normalization.}
A normalizer $N$ maps a document to a cleaned copy.
Following the Unicode standards~\cite{uax15,uts39} and the rules of robustness work~\cite{conpara,humanizing}, $N$ maps 37 Cyrillic and Greek homoglyphs and 7 typographic punctuation marks to their basic Latin forms.
It then applies NFKC and deletes Unicode format characters (category Cf), which include zero-width spaces.
The normalized score $z(N(x))$ holds the guard in our study (Section~\ref{sec:main}).
\sys uses it as its anchor.

\subsection{Motivation}
\label{sec:motiv}

We measure three detectors on the development set of Section~\ref{sec:setup}, with the threshold calibrated at $\alpha=0.1\%$ (Figure~\ref{fig:motiv}).
This set holds 600 machine-generated documents from the RAID benchmark~\cite{raid} spread over eleven attacks and unedited text.

\paragraph{Invisible edits break the score.}
Homoglyph and zero-width edits leave the text unchanged to a reader but split it into many more tokens.
A homoglyph document has 489 tokens on average against 327 after normalization, and a zero-width document has 506 against 279.
The per-token likelihoods of both models shift, and the mean Binoculars ratio rises from 0.80 to 0.93 and 0.85, above the calibrated cut-off near 0.80.
Recall falls from 75.5\% on unedited documents to 4.2\% and 2.0\% (Figure~\ref{fig:motiv}a).
We call this \shatter.
\emph{Edits that a reader cannot see move machine text into the human range of the score,} which motivates normalized scoring (Section~\ref{sec:anchor}).

\paragraph{Cleaning repairs characters only.}
Normalization restores recall to 56.3\% and 62.0\% on the two edited slices and leaves recall on the other ten slices unchanged.
It adds no false positive on the 7,521 human documents of our guard set, with 7 in both cases.
The documents it still misses come from attacks with nothing left to clean.
Recall stays at 14.5\% under synonym swaps and 45.9\% under paraphrasing (Figure~\ref{fig:motiv}b).
An unconditional mapping table also rewrites characters in human URLs, code and names, so any extra evidence must be checked against human text of every section.
\emph{Normalization closes the character gap, and the remaining half of machine text needs evidence beyond the normalized score,} which motivates adding a learned head (Section~\ref{sec:head}).

\paragraph{Learned heads lose the tail.}
The natural next step is a classifier on frozen token statistics of the normalized document.
At $\alpha=0.1\%$, every head we trained alone recalls less than the normalized score, which reaches 50.5\%.
Trained on 1,576 labelled documents, a logistic head over seven statistics reaches 28.8\% with one head for all sections and 47.7\% with one head per section.
With all 5,692 documents of the training pool and the eight statistics of Section~\ref{sec:anchor}, the per-section head reaches 47.3\%.
Heads over 52 distribution features reach 23.2--38.0\% and flag too many human reviews or forum posts (hatched in Figure~\ref{fig:motiv}c).
\emph{On this pair, a head ranks the extreme tail of human text worse than one calibrated zero-shot score.}
This motivates anchoring the head to that score and choosing its weight under a human FPR cap (Sections~\ref{sec:blend} and~\ref{sec:kappa}).

\subsection{Overview}
\label{sec:overview}

Figure~\ref{fig:overview} summarizes \sys.
Normalized scoring and a per-section head feed an anchored blend, which addresses C1 (Sections~\ref{sec:anchor} to~\ref{sec:blend}).
A guarded weight and one threshold calibrated on human text address C2 (Section~\ref{sec:kappa}).

\subsection{Normalized Scoring}
\label{sec:anchor}

\sys scores the normalized copy $N(x)$ of every document, so the language models never see the homoglyphs and format characters that $N$ maps or deletes.
From the same forward passes of $M_1$ and $M_2$ over $N(x)$, it keeps the anchor score $z(N(x))$ and a feature vector $\phi(x)\in\mathbb{R}^8$.
The features are the two terms of Equation~\eqref{eq:bino}, the ratio, the score $z$ and the mean negative log-likelihood under the observer.
The other three are the standard deviations of the per-token negative log-likelihood under the performer and of the per-token cross-entropy, and the length in tokens divided by 512.
They describe how surprising the text is and how evenly the surprise is spread, which the ratio averages away.

\subsection{Per-Section Head}
\label{sec:head}

Human writing differs by section, as recipes are formulaic and forum posts are terse, so one decision rule fits some sections poorly.
\sys therefore fits one head per section $d$:
\begin{equation}
h(x)=\sigma\!\left(\mathbf{w}_{d}^{\top}\,\mathrm{std}_{d}(\phi(x))+b_{d}\right),\quad d=d(x),
\label{eq:head}
\end{equation}
where $\mathrm{std}_d$ standardizes each feature with the mean and standard deviation of section $d$ in the training pool, and $\sigma$ is the logistic function.
Each head is an $\ell_2$-regularized logistic regression with balanced class weights.
The training pool holds 5,692 labelled documents from the RAID training split whose sources are disjoint from every calibration, development and guard document.
Its machine documents follow the same stratification over generators, decoding settings and attacks as the development set.
The section of a document is known at serving time on a Web platform, since every post arrives through a forum, a review page or a comment box.

\subsection{Anchored Blend}
\label{sec:blend}

\sys keeps the normalized score as an anchor and adds the head as a correction:
\begin{equation}
s(x)=(1-\kappa)\,\frac{z(N(x))-\mu_z}{\sigma_z}+\kappa\,\frac{h(x)-\mu_h}{\sigma_h},
\label{eq:blend}
\end{equation}
where $\mu$ and $\sigma$ are the mean and standard deviation of each score over the training pool, and $\kappa\in[0,1]$ is the weight of the head.
With $\kappa=0$ the blend ranks documents as the normalized score does.
Both terms are standardized, so $\kappa$ sets how far the head can move a document away from its anchor position.
The anchor keeps the order of the extreme tail, and the head lifts machine documents that sit just below the threshold.

\subsection{Guarded Weight and Calibration}
\label{sec:kappa}

The weight $\kappa$ decides how far the head may move the tail, and choosing it on evaluation data would leak.
\sys chooses it with five folds of the training pool, split by a hash of the source identifier so that documents of one source stay in one fold.
For every $\kappa$ in a grid $\mathcal{K}=\{0, 0.02, 0.05, 0.1, 0.2, 0.4\}$ and every fold, it fits the heads on the other four folds and sets the threshold on $\mathcal{H}_{\text{cal}}$ at $\alpha$.
It then reads the TPR and the human FPR on the held-out fold.
It keeps the weights whose human FPR stays at most 0.5\% in every fold.
Among them it picks the one with the highest mean TPR, and the smaller one on a tie:
\begin{equation}
\kappa^{*}=\arg\max_{\kappa\in\mathcal{K}}\ \overline{\mathrm{TPR}}(\kappa)\quad\text{s.t.}\quad \max_{f}\mathrm{FPR}_{f}(\kappa)\le 0.5\%.
\label{eq:kappa}
\end{equation}
The cap is half of the guard, which leaves room for the shift between training and deployment.
Because $\kappa=0$ is in the grid, the selection can fall back to the anchor whenever the anchor meets the cap.
After selection, \sys refits the heads on the full pool and sets one global threshold $\tau$ on $\mathcal{H}_{\text{cal}}$ with the RAID threshold finder~\cite{raid}.
On our pool the rule selects $\kappa^{*}=0.4$ for the Falcon pair.

\paragraph{Cost.}
\sys runs the same two forward passes as Binoculars on one view of the document.
The features come from the logits these passes already produce, and a head adds one dot product of length eight.
With the Falcon-7B pair, a document takes 0.35~s at a peak of 26.2~GiB, on par with Binoculars and its normalized version (Table~\ref{tab:cost}).
The text normalization takes 0.8~ms of it on the CPU.

%======================================================================
\section{Experiments}
\label{sec:eval}
%======================================================================

\begin{figure*}[t]
\centering
\begin{subfigure}[t]{0.343\linewidth}\includegraphics[width=\linewidth]{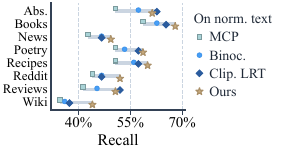}\caption{Recall per section}\end{subfigure}%
\begin{subfigure}[t]{0.285\linewidth}\includegraphics[width=\linewidth]{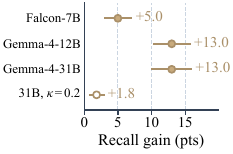}\caption{Gain per scoring pair}\end{subfigure}%
\begin{subfigure}[t]{0.361\linewidth}\includegraphics[width=\linewidth]{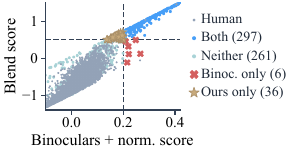}\caption{Blend against normalized score}\end{subfigure}
\Description{Three panels. Recall in each of the eight RAID sections for MCP, normalized Binoculars, the normalized clipped LRT and NormDetect; recall gain of NormDetect over normalized Binoculars with 95\% intervals for the Falcon-7B, Gemma-4-12B and Gemma-4-31B pairs and for Gemma-4-31B at blend weight 0.2; scatter of the normalized Binoculars score against the blend score with both calibrated thresholds dashed.}
\caption{Recall per section, recall gain per scoring pair, and the two scores of each document at a 0.1\% calibrated FPR.}
\label{fig:trio}
\end{figure*}

\subsection{Setup}
\label{sec:setup}

\paragraph{Data.}
We use the public training split of RAID~\cite{raid}, which covers eight Web sections, eleven generators, four decoding settings and eleven attacks.
The sections (domains in RAID) are abstracts, books, news, poetry, recipes, Reddit, reviews and Wikipedia.
Its test split has no public labels, so all sets below come from the training split and are disjoint by source article.
The calibration pool $\mathcal{H}_{\text{cal}}$ holds 3,000 human documents, 375 per section.
The development set holds 600 machine-generated documents stratified over generators, decoding settings and attacks, with twelve attack slices including unedited text.
The guard set $\mathcal{H}_{\text{guard}}$ holds 7,521 human documents, 1,000 per section and 521 for reviews, where fewer unused sources remain.
The training pool of Section~\ref{sec:head} holds 5,692 documents.

\input{tables/tab_main}

\input{tables/tab_cost}

\paragraph{Models.}
The main scoring pair is the one of the original Binoculars paper~\cite{binoculars}, Falcon-7B as observer and Falcon-7B-instruct as performer~\cite{falcon}.
We also swap in the base and instruct models of Gemma-4-12B (Table~\ref{tab:main}) and of Gemma-4-31B~\cite{gemma4} as scoring pairs.
For these pairs we extend the grid $\mathcal{K}$ with 0.6, 0.8 and 1.0, and the rule selects $\kappa^{*}=1.0$ on both.
Appendix~\ref{app:protocol} gives the precision, the token limit and the scoring models of each baseline.

\paragraph{Baselines.}
We compare seventeen detectors (Table~\ref{tab:main}).
Training-free ones are Likelihood and Log-Rank~\cite{gltr}, LRR~\cite{detectllm}, Lastde~\cite{lastde}, Telescope~\cite{telescope}, DNA-GPT~\cite{dnagpt}, Fast-DetectGPT~\cite{fastdetectgpt}, Binoculars~\cite{binoculars} on the original and on the normalized text, and, on the normalized text, the clipped likelihood-ratio test (LRT)~\cite{clippedlrt}.
MCP~\cite{mcp} calibrates the normalized Binoculars score with conformal thresholds per length bin.
Trained ones are RADAR~\cite{radar}, AdaDetectGPT~\cite{adadetectgpt}, a fine-tuned DeBERTa-v3~\cite{debertav3} and adapted versions of DivEye~\cite{diveye}, BiScope~\cite{biscope} and Luminol-AIDetect~\cite{luminol}.
Where a baseline runs on both the original and the normalized text, Table~\ref{tab:main} keeps the view that recalls more.
Appendix~\ref{app:adapt} gives every adaptation, and its Table~\ref{tab:views} lists the recall and guard of each baseline on both views.

\paragraph{Metrics and statistics.}
Every detector gets one global threshold from $\mathcal{H}_{\text{cal}}$ at $\alpha=0.1\%$ with the threshold finder of RAID.
We report the TPR on the development set and check the guard of Section~\ref{sec:pre} on $\mathcal{H}_{\text{guard}}$.
Table~\ref{tab:main} gives the largest per-section bound of each detector, and a cross marks a bound above 1\%.
Bold and underline mark the best and second best detectors of each block whose largest bound is at most 1\%.
The Edited column covers the homoglyph and zero-width documents.
Differences in TPR come with 95\% confidence intervals from a paired bootstrap~\cite{efron79} with 10,000 resamples over the 574 source articles of the development set.
Table~\ref{tab:cost} reports the median time of three runs on one A800 GPU.

\subsection{Main Results}
\label{sec:main}

\begin{figure*}[t]
\centering
\includegraphics[width=\textwidth]{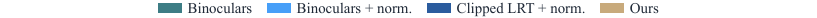}\\
\includegraphics[width=0.506\linewidth]{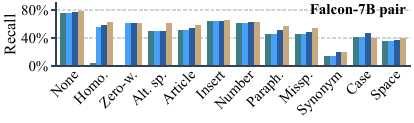}%
\includegraphics[width=0.483\linewidth]{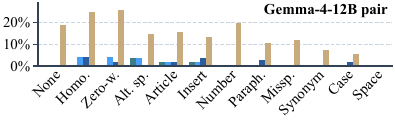}
\Description{Two grouped bar charts of recall on twelve attack slices for Binoculars, normalized Binoculars, the normalized clipped LRT and NormDetect, one for the Falcon-7B pair and one for the Gemma-4-12B pair.}
\caption{Recall per attack slice at a 0.1\% calibrated FPR with two scoring pairs.}
\label{fig:slices}
\end{figure*}

\paragraph{\sys leads every guarded detector.}
At the main operating point with the Falcon pair, \sys recalls 55.5\% of machine-generated documents and holds the guard in every section (Table~\ref{tab:main}).
It gains 5.0 points over normalized Binoculars and 2.3 points over the clipped LRT.
Among guarded detectors it also leads on homoglyphs and ties on zero-width edits (Figure~\ref{fig:slices}, Falcon-7B pair).

\paragraph{Both gains are significant.}
Under the paired bootstrap over source articles, the gain over normalized Binoculars has an interval of [2.98, 7.15] points.
The interval of the gain over the clipped LRT starts at 0.33 points, so neither interval includes zero.
The heads are deterministic, and three training seeds give identical decisions.

\begin{figure*}[t]
\centering
\includegraphics[width=\textwidth]{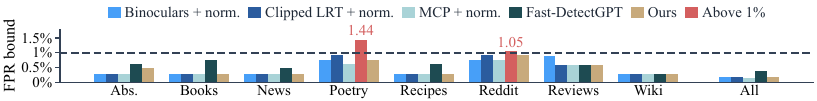}
\Description{Grouped bar chart of the one-sided 95\% upper bound of the human FPR in each of the eight guard sections and in the pooled guard set for normalized Binoculars, the normalized clipped LRT, MCP, Fast-DetectGPT and NormDetect, with a dashed line at 1\%. Only the Fast-DetectGPT bars on poetry and Reddit cross the line.}
\caption{Upper bound of the human FPR per guard section at a 0.1\% calibrated FPR.}
\label{fig:guard}
\end{figure*}

\paragraph{Higher recall breaks the guard.}
Fast-DetectGPT recalls more at the nominal threshold but flags 21 human documents, nearly three times the calibrated rate, and its bounds exceed 1\% on poetry and Reddit (Figure~\ref{fig:guard}).
Table~\ref{tab:guard} in Appendix~\ref{app:guard} counts the human flags of every detector per guard section.
It shows that the six trained detectors also break the guard, each in one to three sections.
When Appendix~\ref{app:fdgpt} moves the threshold of \sys on the guard set to the same 21 human flags, the two detectors recall within one point of each other.

\paragraph{Every section stays under 1\%.}
\sys flags one more guard document than normalized Binoculars, and every per-section bound stays below the guard (Figure~\ref{fig:guard}).
Its largest bound is 0.91\% on Reddit posts, level with the clipped LRT, so no section pays for the gain.
Over the pooled guard set, its eight false positives give a bound of 0.19\%, about a fifth of the guard.

\paragraph{Gains reach every section.}
\sys recalls more than normalized Binoculars in all eight sections, with 75 development documents each (Figure~\ref{fig:trio}a).
The largest gain is 8.0 points on Wikipedia, where every detector recalls the least.
\sys also recalls the most among the guarded detectors in six of the eight sections.
On Reddit posts, where normalized Binoculars and the clipped LRT recall the same share, it gains 5.3 points.

\begin{figure}[t]
\centering
\includegraphics[width=\linewidth]{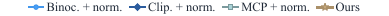}\\
\begin{subfigure}[t]{0.519\linewidth}\includegraphics[width=\linewidth]{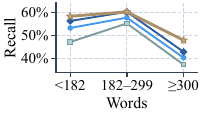}\caption{Document length}\end{subfigure}%
\begin{subfigure}[t]{0.470\linewidth}\includegraphics[width=\linewidth]{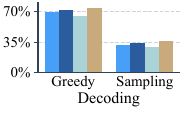}\caption{Decoding}\end{subfigure}
\Description{Left, recall of four detectors on the three length thirds of the development set. Right, recall of the same detectors under greedy decoding and under sampling.}
\caption{Recall per length third and per decoding strategy of the development set at a 0.1\% calibrated FPR.}
\label{fig:lendec}
\end{figure}

\paragraph{Gains hold across length and decoding.}
We split the development set into thirds by word count and by decoding strategy (Figure~\ref{fig:lendec}).
Among the four detectors there, \sys recalls the most or ties in every third.
On the longest third, its gain over normalized Binoculars is 7.5 points, the largest of the three thirds.
The shortest third gains as well, so the head also helps on short posts.
Under greedy decoding, \sys gains 5.4 points over normalized Binoculars.
Under sampling, where every detector recalls less than half of the machine text, it gains 4.6 points.

\paragraph{Gains hold on Gemma-4.}
On the Gemma-4-12B pair, \sys gains 13.0 points over normalized Binoculars, its paired bootstrap interval excludes zero, and every section holds the guard (Figure~\ref{fig:trio}b, Table~\ref{tab:main}).
Its recall on edited documents is more than three times that of any baseline.
Table~\ref{tab:g31} in Appendix~\ref{app:g31} compares three blend weights on the 31B pair.
There the selected weight trades the Reddit guard for recall, and $\kappa=0.2$ keeps every section within the guard with a gain of 1.8 points.
On Ministral-3-8B~\cite{ministral3} and Qwen3.5-9B~\cite{qwen35} the gain of \sys vanishes at equal FPR, and pairing the blend with stronger scoring models is left for future work.

\subsection{Attacks}
\label{sec:attacks}

\paragraph{Gains peak where cleaning fails.}
\sys recalls more than normalized Binoculars on most slices and matches it on zero-width edits (Figure~\ref{fig:slices}, Falcon-7B pair).
The largest gains, about ten points each, appear under alternative spelling, paraphrasing and misspelling.
Under paraphrasing, for example, \sys recalls 56.8\% against 45.9\% for normalized Binoculars.
Normalization leaves recall on these slices unchanged, so the head supplies the evidence that cleaning cannot.

\paragraph{Gains carry over to Gemma-4-12B\@.}
\sys leads on every slice where any detector flags a document (Figure~\ref{fig:slices}, Gemma-4-12B pair).
The three baselines stay below 5\% recall on every slice, while \sys reaches 26.0\% on zero-width edits.
On unedited text, where none of the three flags a document, \sys recalls 18.9\% of the machine text.

\begin{figure*}[t]
\centering
\includegraphics[width=\textwidth]{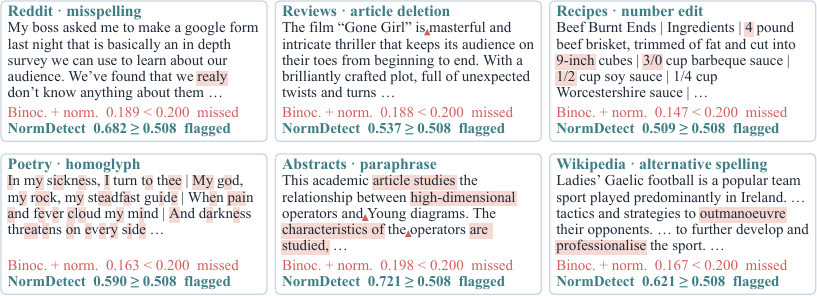}
\Description{Six text cards in two rows. The first row holds Reddit, reviews and recipes posts with a misspelling, a deleted article and changed numbers. The second row holds a poem with homoglyphs, a paraphrased abstract and a Wikipedia article with British spellings. Edited words or letters are shaded. Each card lists the score of normalized Binoculars below its threshold and the score of NormDetect above its threshold.}
\caption{Machine-generated documents that \sys flags and normalized Binoculars misses at a 0.1\% calibrated FPR. Shading marks words or letters the attack changed, and a caret marks deleted words.}
\label{fig:examples}
\end{figure*}

\begin{figure*}[t]
\centering
\begin{subfigure}[t]{0.247\linewidth}\includegraphics[width=\linewidth]{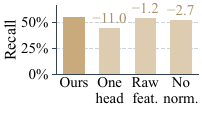}\caption{Components}\end{subfigure}%
\begin{subfigure}[t]{0.252\linewidth}\includegraphics[width=\linewidth]{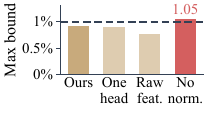}\caption{Largest section bound}\end{subfigure}%
\begin{subfigure}[t]{0.247\linewidth}\includegraphics[width=\linewidth]{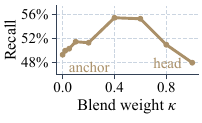}\caption{Blend weight}\end{subfigure}%
\begin{subfigure}[t]{0.243\linewidth}\includegraphics[width=\linewidth]{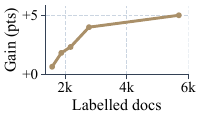}\caption{Training pool}\end{subfigure}
\Description{Four panels. Recall of NormDetect and of three variants with one head for all sections, head features from the original text, and no normalization, with paired differences over the bars. The largest per-section human FPR bound of the same four, with a dashed line at 1\% that only the variant without normalization crosses. Recall of NormDetect against the blend weight, from the anchor alone to the head alone. Recall gain over normalized Binoculars against the number of labelled training documents.}
\caption{Recall and largest per-section human FPR bound of \sys and three variants, recall against the blend weight, and gain against the size of the training pool.}
\label{fig:ablation}
\end{figure*}

\subsection{Examples}
\label{sec:examples}

\paragraph{Edits that cleaning keeps.}
The first row of Figure~\ref{fig:examples} shows three development posts from Reddit, reviews and recipes.
Their misspelling, missing article and changed number all survive normalization.
Normalized Binoculars scores each post below its threshold, and \sys flags all three.
\sys scores the misspelled Reddit post at 0.682, the highest in the row.

\paragraph{Some misses need the head.}
The second row shows a poem, an abstract and a Wikipedia article.
Normalization restores every homoglyph in the poem, yet normalized Binoculars still scores it below its threshold.
The paraphrase and the British spellings leave nothing to clean, and \sys flags all three documents.
The six cards span five generators and both decoding strategies.
Across them, \sys clears its threshold of 0.508, while normalized Binoculars stays below its threshold of 0.200.

\subsection{Ablation}
\label{sec:ablation}

\paragraph{Per-section heads matter most.}
With one head for all sections at the same weight, recall drops by 11.0 points (Figure~\ref{fig:ablation}a).
The guarded choice of $\kappa$ then gives that head zero weight, which leaves the anchor alone at 49.3\%.
The one-head variant still holds the guard, so its loss lies in recall alone.
Reading the head features from the original text costs only 1.2 points and also holds the guard.

\paragraph{Normalization also protects the guard.}
Without normalization of both the anchor and the head, recall falls by 2.7 points (Figure~\ref{fig:ablation}a).
The Reddit bound then rises to 1.05\%, above the guard (Figure~\ref{fig:ablation}b).
It also flags four more guard documents than \sys.

\paragraph{Both terms of the blend matter.}
The anchor alone and the head alone each recall about half of the machine text (Figure~\ref{fig:ablation}c).
Blending them lifts recall to 55.5\%, and the next larger weight gives nearly the same recall.
Weights above it lose recall, down to 48.0\% for the head alone.
Protecting URLs, code and numbers from normalization leaves every recall and false positive count unchanged.

\paragraph{The head lifts near misses.}
Against normalized Binoculars, \sys flags 36 development documents that the anchor misses and drops 6 that it flags (Figure~\ref{fig:trio}c).
All of them lie within 0.07 of the normalized threshold, and misspelling and alternative spelling are the most common attacks among them.
The head thus refines the anchor near its threshold rather than replacing it.
On the guard set the blend adds three human flags and removes two.

\paragraph{More labels help steadily.}
The gain over normalized Binoculars grows steadily with the training pool, from under one point with the smallest pool to 5.0 points with the full pool (Figure~\ref{fig:ablation}d).
Adding 1,200 labelled documents to the smallest pool already lifts the gain to 4.0 points.

\subsection{Robustness}
\label{sec:robust}

\begin{figure*}[t]
\centering
\includegraphics[width=\linewidth]{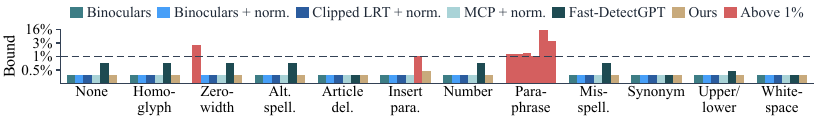}
\caption{Human FPR bound when each RAID attack edits the same 1,000 human documents, at the thresholds of Table~\ref{tab:main}.}
\Description{Grouped bars for six detectors over twelve groups, the unedited documents and the eleven RAID attacks, with a dashed line at a 1 percent bound. On the unedited documents and on the ten edits other than paraphrase, NormDetect stays below the line. On paraphrase every detector is above the line, and Fast-DetectGPT reaches 15.8 percent.}
\label{fig:edited}
\end{figure*}

\begin{figure}[t]
\centering
\begin{subfigure}[t]{0.517\linewidth}\includegraphics[width=\linewidth]{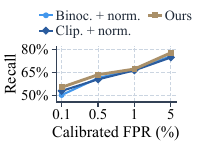}\caption{Calibrated targets}\end{subfigure}%
\begin{subfigure}[t]{0.472\linewidth}\includegraphics[width=\linewidth]{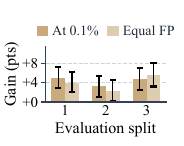}\caption{Gain over Binoc. + norm.}\end{subfigure}
\Description{Left, recall against the calibrated FPR target for three detectors. Right, recall gain of NormDetect over normalized Binoculars with 95\% intervals on three evaluation splits, at the calibrated threshold and at equal human false positives.}
\caption{Recall at four calibrated FPR targets and the gain of \sys on three evaluation splits.}
\label{fig:points}
\end{figure}

\paragraph{Wording-preserving edits hold the bound.}
We apply each of the eleven RAID attacks to the same 1,000 human documents of the guard set, spread evenly over the eight sections, and keep the thresholds of Table~\ref{tab:main}.
On the ten edits that keep the writer's wording, the pooled human FPR bound of \sys stays at most 1\%, as Figure~\ref{fig:edited} shows on a piecewise axis.
\sys flags no edited document under nine of these edits and one under paragraph insertion.
Without normalization, Binoculars exceeds the bound on these human documents under zero-width spaces.
Normalization thus also protects human writers whose text carries invisible characters.
Paraphrase by DIPPER~\cite{dipper} rewrites the whole text in language-model wording, and every detector shown exceeds 1\% on it.

\paragraph{Gains hold on redrawn splits.}
We redraw the calibration and development sets twice by hashing source articles with two other seeds.
\sys gains 3.2 and 4.7 points over normalized Binoculars, and both intervals exclude zero.
On both splits, its human FPR bound stays at most 1\% in every section.
The gains stay positive, and their intervals exclude zero, when both detectors flag equally many human texts (Figure~\ref{fig:points}b).

\subsection{Operating Points}
\label{sec:points}

\begin{figure}[t]
\centering
\includegraphics[width=\linewidth]{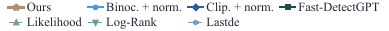}\\
\begin{subfigure}[t]{0.518\linewidth}\includegraphics[width=\linewidth]{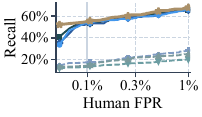}\caption{Falcon-7B pair}\end{subfigure}%
\begin{subfigure}[t]{0.471\linewidth}\includegraphics[width=\linewidth]{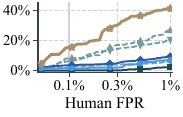}\caption{Gemma-4-12B pair}\end{subfigure}
\Description{Two line charts of recall against the human FPR on the guard set, on a log axis from about 0.05\% to 1\%, for seven detectors with the Falcon-7B pair and with the Gemma-4-12B pair. A light vertical line marks 0.1\%.}
\caption{Recall against the human FPR when the threshold is swept over the guard documents, with two scoring pairs.}
\label{fig:lowfpr}
\end{figure}

\paragraph{The lead is widest at 0.1\%.}
\sys stays above normalized Binoculars and the clipped LRT at all four calibrated targets (Figure~\ref{fig:points}).
Its lead over normalized Binoculars is widest at 0.1\%, the only target where any detector holds the guard.
There it is 5.0 points, more than twice its lead at any looser target.
At the loosest target, \sys still recalls the most of the three.

\paragraph{The lead holds without calibration.}
Here we sweep the threshold over the guard documents themselves, with no calibration set (Figure~\ref{fig:lowfpr}).
Across the plotted range from 0.05\% to 1\% human FPR, \sys recalls the most at every point on both scoring pairs.
On the Falcon pair its lead is largest at the strict end, 11.0 points over Fast-DetectGPT.

\paragraph{Baselines stay low on Gemma-4-12B\@.}
On the Gemma-4-12B pair, normalized Binoculars and Fast-DetectGPT stay below a tenth of the machine text across the whole range (Figure~\ref{fig:lowfpr}b).
In Table~\ref{tab:main}, Likelihood is the strongest baseline on this pair.
At 1\% human FPR, \sys recalls 41.3\%, which is 14.5 points above Likelihood at the same rate.

\paragraph{Gains hold at equal false positives.}
\sys flags one more human document of the guard set than normalized Binoculars.
With its threshold moved to the same false positive count, \sys still recalls 4.0 points more, and the interval excludes zero.
The gain therefore does not come from the extra human flag.
On the Gemma-4-12B pair, the gain at equal false positives is 9.5 points, and its interval also excludes zero.

%======================================================================
\section{Conclusion}
\label{sec:conclusion}
%======================================================================
\sys detects machine-generated Web text by anchoring a learned per-section head to the normalized Binoculars score, as motivated by \shatter and by the weak tail of learned heads.
Normalization removes the invisible edits, the head adds evidence where nothing is left to clean, and a weight chosen under a human FPR cap keeps the blend safe for every section.
On eight RAID sections with the main scoring pair, \sys leads normalized Binoculars by 5.0 points at a 0.1\% calibrated FPR and recalls the most among detectors that hold the guard.
With the Gemma-4-12B pair, the gain grows to 13.0 points, and \sys recalls more than three times as much as any baseline on that pair while holding the guard.
\sys adds no forward pass to Binoculars, only one small head per section.
We expect the same anchoring to help other detectors where a single false flag hurts a human writer.

\bibliographystyle{ACM-Reference-Format}
\bibliography{refs_gen,refs_manual}

\appendix

\section{Human False Positives per Section}
\label{app:guard}
Table~\ref{tab:guard} lists the human false positives of each detector in every section of the guard set.
A star marks a bound above 1\%.
\input{tables/tab_guard}

\section{Fast-DetectGPT at Its Own False Positive Rate}
\label{app:fdgpt}
Fast-DetectGPT reaches its nominal recall in Table~\ref{tab:main} while flagging 21 of the 7,521 guard documents, which is 2.8 times the calibrated rate.
As a diagnostic, we move the threshold of \sys on the guard set until it flags the same 21 documents.
\sys then recalls 61.2\% against 60.3\% for Fast-DetectGPT, a paired difference of 0.8 points.
Its interval includes zero, so the two are comparable at this rate.
This threshold is read on evaluation data and is not an operating point of either detector.

\section{The Gemma-4-31B Pair}
\label{app:g31}
Table~\ref{tab:g31} lists normalized Binoculars and \sys on the Gemma-4-31B pair for three blend weights, with the full training pool.
The rule selects $\kappa^{*}=1.0$, where the Reddit bound reaches 1.05\%.
A cross marks a detector with a section bound above 1\%.
\input{tables/tab_g31}

\section{Protocol Details}
\label{app:protocol}
Both models of each scoring pair run with Transformers~\cite{transformers} in bfloat16 precision with at most 512 tokens per document.
Single-model statistics use the instruct model of each pair on both the original and the normalized text.
On the 12B pair we also rerun the training-free baselines that use it.
DNA-GPT and the trained detectors other than AdaDetectGPT use their own models, and AdaDetectGPT runs with the Falcon pair, so Table~\ref{tab:main} lists them only once.
Fast-DetectGPT runs on the normalized text with the same Falcon pair as \sys.

The threshold finder of RAID steps from the mean human score and stops at the first threshold that meets $\alpha$.
Scores with the same ranking can thus recall slightly different shares, such as 49.3\% for the blend at $\kappa=0$ and 50.5\% for normalized Binoculars.
The interval of the gain over the clipped LRT is [0.33, 4.37] points.
Table~\ref{tab:cost} times each detector on the first 200 development documents three times, on one A800 GPU with no other process and the models loaded, and reports the median.
Detectors that read the normalized text are timed with normalization included.

\section{Baseline Adaptations}
\label{app:adapt}
Lastde, DNA-GPT, the clipped LRT, DivEye, BiScope and Luminol-AIDetect are adapted implementations.

\paragraph{Zero-shot statistics.}
Likelihood and Log-Rank use the mean log-likelihood and the mean log-rank of the observed tokens~\cite{gltr}, and LRR divides the two~\cite{detectllm}.
Telescope reads the probability that a token repeats once it is in the context~\cite{telescope}.
Lastde follows Equations 1--9 of its paper with the published settings and orients the score so that machine text scores higher~\cite{lastde}.
On the original text its scores tie at the calibration threshold, so its calibrated FPR there is 0.13\%.
These five run on both views, and Table~\ref{tab:views} lists each view on both scoring pairs.
A cross there marks a detector that breaks the guard.
DivEye and Luminol use their own models and AdaDetectGPT runs with the Falcon pair, so the table lists the three for the Falcon pair only.
Fast-DetectGPT uses the analytic criterion of its official code with the Falcon pair~\cite{fastdetectgpt}.
\input{tables/tab_views}
\paragraph{DNA-GPT}
We truncate each text at half its length and regenerate five continuations with GPT-2-XL at temperature 0.7 and up to 300 tokens.
We then score the n-gram overlap with the weighting of the paper~\cite{dnagpt}.
Generation failed for 3 calibration documents, which are left out of the threshold.
\paragraph{Clipped LRT}
Following the clipped LRT~\cite{clippedlrt}, we clip the per-token surprise, here in both terms of the Binoculars ratio and at the 95th percentile of the human calibration tokens.
With no clipping the score equals the Binoculars score.
\paragraph{MCP}
MCP takes one conformal quantile of the human calibration scores per bin of token length and flags a document above the quantile of its bin~\cite{mcp}.
We sweep the bin width over 50, 100, 200 and 512 tokens and report the best, a single bin, which recalls 46.7\%.
At the default width of 100 tokens most bins hold fewer than the 999 human documents that a 0.1\% level needs, so they never flag.
\paragraph{Trained detectors.}
RADAR uses the released detector and the label order of its official example~\cite{radar}.
DivEye computes the surprisal statistics of the paper and fits a logistic head on our training pool instead of the original gradient-boosted trees~\cite{diveye}.
Its main row scores the normalized text.
BiScope keeps its bidirectional cross-entropy features, computed with the Falcon pair, and fits a logistic head on our training pool~\cite{biscope}.
Luminol-AIDetect has no released code, so we follow Section~3 of its paper.
GPT-Neo-2.7B~\cite{gptneo} scores each text and one shuffle of it with a sliding window of 2,048 tokens and a stride of 1,024.
A paragraph with several sentences has its sentences shuffled, and a one-sentence paragraph has its words shuffled.
Each of five features of the two perplexities gets one density per section and class, taken from the family with the best Kolmogorov--Smirnov fit among the paper's eleven families.
We make two changes. We fit the densities on our training pool instead of the paper's RAID subset of GPT-3 text, and we score a document by the mean log density ratio over the five features, whose sign is the vote of each feature.
This score takes one global threshold like every other baseline.
The original majority vote flags 14.2\% of the human calibration documents, so it cannot run at 0.1\%.
Its row in Table~\ref{tab:main} uses the original text, which recalls more than the normalized text (Table~\ref{tab:views}).
AdaDetectGPT runs its official code with the Falcon pair of Fast-DetectGPT and learns one witness function on our training pool~\cite{adadetectgpt}.
Its training leaves out the one single-token machine document, and its row uses the normalized text, which recalls more (Table~\ref{tab:views}).

\paragraph{DeBERTa-v3}
DeBERTa-v3 fine-tunes DeBERTa-v3-large with a binary head on our training pool, following the recipe of ConPara~\cite{conpara}.
It reads the first 512 tokens of each text.
It uses AdamW with weight decay 0.01, a peak learning rate of $2\times10^{-5}$, a one-cycle cosine schedule and an effective batch of 32.
Of at most five epochs it keeps the one with the best balanced accuracy on a held-out fold of the training pool.
It then retrains on the whole pool for that many epochs.
Each seed trains once on the original text, and the normalized view feeds the normalized text to the same model.
It scores the difference of the two logits, and its table rows take the median of three seeds.

\section{Ethics}
\label{app:ethics}
All attacks come from the public RAID benchmark~\cite{raid}, and we design no new evasion edits.
The attacks serve only an offline evaluation of detectors.
A detector can still flag human writers, so a platform should send each flag to human review.

\section{Use of Generative Artificial Intelligence}
\label{app:genai}
Generative artificial intelligence assistants helped draft and edit the text and write the experiment and plotting code.
An image-generation model sketched candidate layouts for Figure~\ref{fig:overview}, which we redrew as a vector diagram.
The authors checked every statement, number and figure against the experiment records and take full responsibility for the paper.

\end{document}

%% file: tables/tab_main.tex
\begin{table}[t]
\centering
\small
\setlength{\tabcolsep}{3pt}
\caption{Recall (\%) at a 0.1\% calibrated FPR on RAID per scoring pair. Single-model baselines take the better of the original and normalized text.}
\label{tab:main}
\begin{tabular*}{\linewidth}{@{}c@{\hspace{3pt}}l@{\extracolsep{\fill}}rrrc@{}}
\toprule
 & Method & TPR & Edited & Human FP & Max bound \\
\midrule
\multicolumn{6}{c}{\textit{Falcon-7B pair}} \\
\midrule
 & Likelihood~\cite{gltr} & 14.500 & 22.45 & 8/7521 & 1.05\,\xmark \\
 & Log-Rank~\cite{gltr} & 13.000 & 12.24 & 5/7521 & 0.77 \\
 & LRR~\cite{detectllm} & 8.833 & 9.18 & 5/7521 & 1.05\,\xmark \\
 & Lastde~\cite{lastde} & 12.333 & 13.27 & 3/7521 & 0.91 \\
 & Telescope~\cite{telescope} & 0.167 & 0.00 & 11/7521 & 1.31\,\xmark \\
 & DNA-GPT~\cite{dnagpt} & 3.000 & 1.02 & 6/7521 & 0.77 \\
 & Fast-DetectGPT~\cite{fastdetectgpt} & 60.333 & 62.24 & 21/7521 & 1.44\,\xmark \\
 & Binoculars~\cite{binoculars} & 41.333 & 3.06 & 7/7521 & 0.91 \\
 & Binoculars + norm.~\cite{binoculars} & 50.500 & 59.18 & 7/7521 & 0.91 \\
 & Clipped LRT + norm.~\cite{clippedlrt} & \underline{53.167} & \underline{60.20} & 8/7521 & 0.91 \\
\multirow{-11}{*}{\rotatebox[origin=c]{90}{Training-free}} & MCP + norm.~\cite{mcp} & 46.667 & 57.14 & 5/7521 & 0.77 \\
\midrule
 & RADAR~\cite{radar} & 0.333 & 0.00 & 2/7521 & 1.20\,\xmark \\
 & DivEye~\cite{diveye} & 11.000 & 10.20 & 12/7521 & 1.44\,\xmark \\
 & BiScope~\cite{biscope} & 33.167 & 32.65 & 13/7521 & 1.48\,\xmark \\
 & Luminol~\cite{luminol} & 12.167 & 20.41 & 26/7521 & 2.01\,\xmark \\
 & AdaDetectGPT~\cite{adadetectgpt} & 56.333 & 59.18 & 16/7521 & 1.57\,\xmark \\
 & DeBERTa-v3~\cite{debertav3} & 82.833 & 98.98 & 7/7521 & 1.20\,\xmark \\
\cmidrule(l){2-6}
\multirow{-7}{*}{\rotatebox[origin=c]{90}{Trained}} & \textbf{\sys (ours)} & \textbf{55.500} & \textbf{62.24} & 8/7521 & 0.91 \\
\midrule
\multicolumn{6}{c}{\textit{Gemma-4-12B pair}} \\
\midrule
 & Likelihood~\cite{gltr} & \underline{4.500} & \underline{7.14} & 6/7521 & 0.77 \\
 & Log-Rank~\cite{gltr} & 3.333 & 4.08 & 6/7521 & 0.77 \\
 & LRR~\cite{detectllm} & 3.000 & \underline{7.14} & 5/7521 & 0.91 \\
 & Lastde~\cite{lastde} & 1.167 & 1.02 & 2/7521 & 0.57 \\
 & Telescope~\cite{telescope} & 0.333 & 1.02 & 20/7521 & 2.06\,\xmark \\
 & Fast-DetectGPT~\cite{fastdetectgpt} & 0.167 & 0.00 & 14/7521 & 1.94\,\xmark \\
 & Binoculars~\cite{binoculars} & 0.667 & 0.00 & 6/7521 & 0.63 \\
 & Binoculars + norm.~\cite{binoculars} & 1.333 & 4.08 & 6/7521 & 0.63 \\
 & Clipped LRT + norm.~\cite{clippedlrt} & 1.333 & 3.06 & 3/7521 & 0.57 \\
\multirow{-10}{*}{\rotatebox[origin=c]{90}{Training-free}} & MCP + norm.~\cite{mcp} & 0.333 & 1.02 & 2/7521 & 0.57 \\
\cmidrule(l){2-6}
 & \textbf{\sys (ours)} & \textbf{14.333} & \textbf{25.51} & 7/7521 & 0.91 \\
\bottomrule
\end{tabular*}
\end{table}

%% file: tables/tab_cost.tex
\begin{table}[t]
\centering
\small
\setlength{\tabcolsep}{3pt}
\caption{Time and peak GPU memory per document of every detector in Table~\ref{tab:main} with the Falcon-7B pair on one A800 GPU.}
\label{tab:cost}
\begin{tabular*}{\linewidth}{@{}l@{\extracolsep{\fill}}rrlrr@{}}
\toprule
Detector & Time (s) & GiB & Detector & Time (s) & GiB \\
\midrule
Likelihood & 0.06 & 13.2 & Binoculars + norm. & 0.35 & 26.2 \\
Log-Rank & 0.05 & 13.4 & Clipped LRT + norm. & 0.35 & 26.2 \\
LRR & 0.05 & 13.4 & MCP + norm. & 0.35 & 26.2 \\
Lastde & 0.06 & 13.2 & RADAR & 0.02 & 2.0 \\
Telescope & 0.06 & 13.3 & DivEye & 0.06 & 13.2 \\
DNA-GPT & 0.94 & 8.9 & BiScope & 0.35 & 26.4 \\
Fast-DetectGPT & 0.35 & 26.2 & Luminol & 0.06 & 5.3 \\
Binoculars & 0.37 & 26.2 & \textbf{\sys} & \textbf{0.35} & \textbf{26.2} \\
\bottomrule
\end{tabular*}
\end{table}

%% file: tables/tab_guard.tex
\begin{table*}[t]
\centering
\small
\caption{Human false positives per guard section at a 0.1\% calibrated FPR, with one-sided 95\% upper bounds (\%) in brackets.}
\label{tab:guard}
\begin{tabular*}{\textwidth}{@{}l@{\extracolsep{\fill}}ccccccccc@{}}
\toprule
Detector & Abs. & Books & News & Poetry & Recipes & Reddit & Reviews & Wiki & All \\
$n$ & 1000 & 1000 & 1000 & 1000 & 1000 & 1000 & 521 & 1000 & 7521 \\
\midrule
Binoculars & 0 (0.30) & 0 (0.30) & 0 (0.30) & 3 (0.77) & 0 (0.30) & 3 (0.77) & 1 (0.91) & 0 (0.30) & 7 (0.17) \\
Binoculars + norm. & 0 (0.30) & 0 (0.30) & 0 (0.30) & 3 (0.77) & 0 (0.30) & 3 (0.77) & 1 (0.91) & 0 (0.30) & 7 (0.17) \\
Clipped LRT + norm. & 0 (0.30) & 0 (0.30) & 0 (0.30) & 4 (0.91) & 0 (0.30) & 4 (0.91) & 0 (0.57) & 0 (0.30) & 8 (0.19) \\
MCP + norm. & 0 (0.30) & 0 (0.30) & 0 (0.30) & 2 (0.63) & 0 (0.30) & 3 (0.77) & 0 (0.57) & 0 (0.30) & 5 (0.14) \\
Fast-DetectGPT & 2 (0.63) & 3 (0.77) & 1 (0.47) & 8 (1.44)$^{\ast}$ & 2 (0.63) & 5 (1.05)$^{\ast}$ & 0 (0.57) & 0 (0.30) & 21 (0.40) \\
RADAR & 0 (0.30) & 0 (0.30) & 0 (0.30) & 0 (0.30) & 0 (0.30) & 0 (0.30) & 2 (1.20)$^{\ast}$ & 0 (0.30) & 2 (0.08) \\
DivEye & 0 (0.30) & 0 (0.30) & 0 (0.30) & 0 (0.30) & 8 (1.44)$^{\ast}$ & 4 (0.91) & 0 (0.57) & 0 (0.30) & 12 (0.26) \\
BiScope & 0 (0.30) & 0 (0.30) & 0 (0.30) & 2 (0.63) & 3 (0.77) & 4 (0.91) & 3 (1.48)$^{\ast}$ & 1 (0.47) & 13 (0.27) \\
Luminol & 2 (0.63) & 10 (1.69)$^{\ast}$ & 0 (0.30) & 0 (0.30) & 3 (0.77) & 6 (1.18)$^{\ast}$ & 5 (2.01)$^{\ast}$ & 0 (0.30) & 26 (0.48) \\
AdaDetectGPT & 2 (0.63) & 1 (0.47) & 0 (0.30) & 9 (1.57)$^{\ast}$ & 1 (0.47) & 3 (0.77) & 0 (0.57) & 0 (0.30) & 16 (0.32) \\
DeBERTa-v3 & 0 (0.30) & 2 (0.63) & 0 (0.30) & 2 (0.63) & 1 (0.47) & 0 (0.30) & 2 (1.20)$^{\ast}$ & 0 (0.30) & 7 (0.17) \\
\midrule
\textbf{\sys} & 1 (0.47) & 0 (0.30) & 0 (0.30) & 3 (0.77) & 0 (0.30) & 4 (0.91) & 0 (0.57) & 0 (0.30) & 8 (0.19) \\
\bottomrule
\end{tabular*}
\end{table*}

%% file: tables/tab_g31.tex
\begin{table}[t]
\centering
\small
\setlength{\tabcolsep}{3pt}
\caption{Recall (\%) at a 0.1\% calibrated FPR on RAID with the Gemma-4-31B pair, by blend weight.}
\label{tab:g31}
\begin{tabular*}{\linewidth}{@{}l@{\extracolsep{\fill}}rcr@{}}
\toprule
Method & TPR & Gain [95\% CI] & Human FP \\
\midrule
Binoculars + norm. & 7.667 & -- & 5/7521 \\
\sys, $\kappa{=}0.2$ & 9.500 & +1.83 [0.67, 3.03] & 6/7521 \\
\sys, $\kappa{=}0.4$ & 14.333 & +6.67 [4.66, 8.84] & 9/7521\,\xmark \\
\sys, $\kappa{=}1.0$ & 20.667 & +13.00 [9.95, 16.06] & 10/7521\,\xmark \\
\bottomrule
\end{tabular*}
\end{table}

%% file: tables/tab_views.tex
\begin{table}[t]
\centering
\small
\caption{Recall (\%) of single-model baselines at a 0.1\% calibrated FPR on the original and the normalized text.}
\label{tab:views}
\begin{tabular*}{\linewidth}{@{}l@{\extracolsep{\fill}}rrrr@{}}
\toprule
 & \multicolumn{2}{c}{Falcon-7B} & \multicolumn{2}{c}{Gemma-4-12B} \\
\cmidrule(lr){2-3}\cmidrule(l){4-5}
Method & Original & Normalized & Original & Normalized \\
\midrule
Likelihood & 14.500\,\xmark & 13.167\,\xmark & 4.500 & 4.167 \\
Log-Rank & 11.667 & 13.000 & 3.333 & 3.333 \\
LRR & 7.667\,\xmark & 8.833\,\xmark & 3.000 & 2.167 \\
Telescope & 0.167\,\xmark & 0.167\,\xmark & 0.167\,\xmark & 0.333\,\xmark \\
Lastde & 9.333 & 12.333 & 1.000 & 1.167 \\
DivEye & 10.500\,\xmark & 11.000\,\xmark & -- & -- \\
Luminol & 12.167\,\xmark & 10.167\,\xmark & -- & -- \\
AdaDetectGPT & 51.500\,\xmark & 56.333\,\xmark & -- & -- \\
DeBERTa-v3 & 82.833\,\xmark & 77.833\,\xmark & -- & -- \\
\bottomrule
\end{tabular*}
\end{table}